\documentclass{article}
\usepackage[preprint]{neurips_2022}
\usepackage[utf8]{inputenc} 
\usepackage[T1]{fontenc}    
\usepackage{hyperref}       
\usepackage{url}            
\usepackage{booktabs}       
\usepackage{amsfonts}       
\usepackage{nicefrac}       
\usepackage{microtype}      
\usepackage{xcolor}         

\newcommand*\samethanks[1][\value{footnote}]{\footnotemark[#1]}
\usepackage{graphicx}
\usepackage{floatrow}
\usepackage{tikz}
\usepackage{comment}
\usepackage{amsmath,amssymb} 
\usepackage{color}
\usepackage{graphicx}
\usepackage{amsmath}
\usepackage{amssymb}
\usepackage{multirow}
\usepackage{booktabs}
\usepackage{adjustbox}
\usepackage{siunitx}
\usepackage{graphicx}
\usepackage{caption}
\usepackage{booktabs, makecell}
\usepackage{amsthm}
\usepackage{bbm}
\usepackage{subcaption}
\usepackage{wrapfig}
\usepackage{natbib}
\setcitestyle{numbers,square}

\title{NUWA-XL: Diffusion over Diffusion \\ for eXtremely Long Video Generation}

\author{Shengming Yin$^{1}$\thanks{Both authors contributed equally to this research.} \quad Chenfei Wu$^{2}$\samethanks[1] \quad Huan Yang$^{2}$ \quad  Jianfeng Wang$^{3}$ \quad Xiaodong Wang$^{2}$\\ \textbf{Minheng Ni}$^{2}$\quad  \textbf{Zhengyuan Yang}$^{3}$\quad  \textbf{Linjie Li}$^{3}$\quad \textbf{Shuguang Liu}$^{2}$\quad \textbf{Fan Yang}$^{2}$  \quad \textbf{Jianlong Fu}$^{2}$ \\ \textbf{Gong Ming}$^{2}$\quad \textbf{Lijuan Wang}$^{3}$\quad \textbf{Zicheng Liu}$^{3}$\quad \textbf{Houqiang Li}$^{1}$\quad \textbf{Nan Duan}$^{2}$\thanks{Corresponding author.} \\
 {$^{1}$University of Science and Technology of China\quad $^{2}$Microsoft Research Asia
 \quad $^{3}$Microsoft Azure AI} \\
 {\tt\small \{sheyin@mail.,lihq@\}ustc.edu.cn, \{chewu,huan.yang,jianfw,v-xiaodwang,t-mni,zhengyang,} \\
{\tt\small lindsey.li,shuguanl,fanyang,jianf,migon,lijuanw,zliu,nanduan\}@microsoft.com }
}

\begin{document}

\maketitle

\begin{abstract}
  In this paper, we propose NUWA-XL, a novel Diffusion over Diffusion architecture for eXtremely Long video generation. Most current work generates long videos segment by segment sequentially, which normally leads to the gap between training on short videos and inferring long videos, and the sequential generation is inefficient. Instead, our approach adopts a ``coarse-to-fine'' process, in which the video can be generated in parallel at the same granularity. A global diffusion model is applied to generate the keyframes across the entire time range, and then local diffusion models recursively fill in the content between nearby frames. This simple yet effective strategy allows us to directly train on long videos (3376 frames) to reduce the training-inference gap, and makes it possible to generate all segments in parallel. To evaluate our model, we build FlintstonesHD dataset, a new benchmark for long video generation. Experiments show that our model not only generates high-quality long videos with both global and local coherence, but also decreases the average inference time from 7.55min to 26s (by 94.26\%) at the same hardware setting when generating 1024 frames. The homepage link is \url{https://msra-nuwa.azurewebsites.net/}
\end{abstract}

\section{Introduction} \label{sec:intro}

\begin{figure*}
    \centering
    \includegraphics[width=\textwidth]{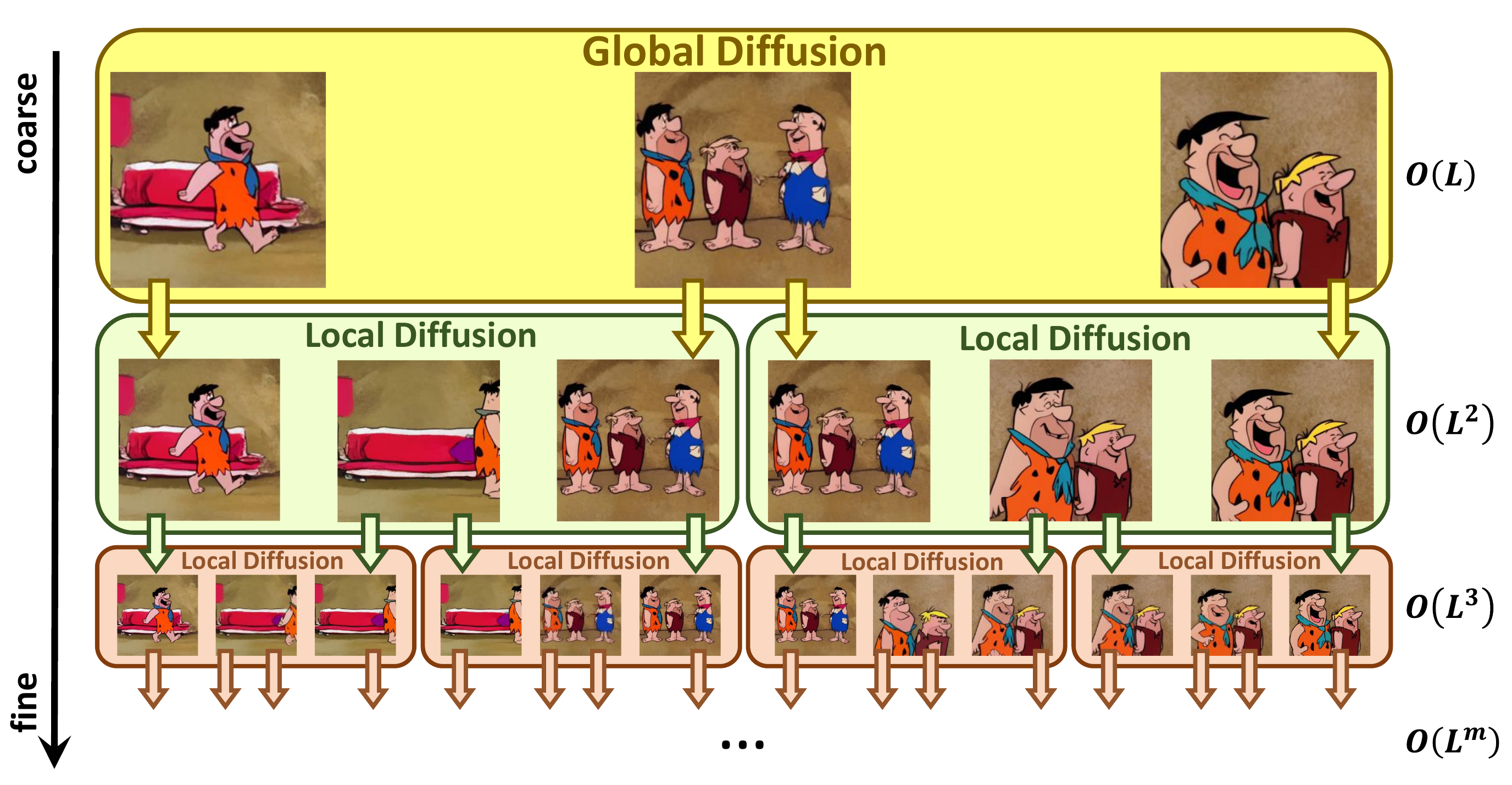}
    \caption{Overview of NUWA-XL for extremely long video generation in a ``coarse-to-fine'' process. A global diffusion model first generates $L$ keyframes which form a ``coarse''  storyline of the video, a series of local diffusion models are then applied to the adjacent frames, treated as the first and the last frames, to iteratively complete the middle frames resulting $O(L^m)$ ``fine''  frames in total.}
    \label{fig:nuwa-xl}
\end{figure*}

Recently, visual synthesis has attracted a great deal of interest in the field of generative models. Existing works have demonstrated the ability to generate high-quality images~\citep{rameshZeroShotTexttoImageGeneration2021,sahariaPhotorealisticTexttoImageDiffusion2022,rombachHighResolutionImageSynthesis2022} and short videos (e.g., 4 seconds~\citep{wuUWAVisualSynthesis2022}, 5 seconds~\citep{singerMakeAVideoTexttoVideoGeneration2022}, 5.3 seconds~\citep{hoImagenVideoHigh2022}). However, videos in real applications are often much longer than 5 seconds. A film typically lasts more than 90 minutes. A cartoon is usually 30 minutes long. Even for ``short'' video applications like TikTok, the recommended video length is 21 to 34 seconds. Longer video generation is becoming increasingly important as the demand for engaging visual content continues to grow. 

However, scaling to generate long videos has a significant challenge as it requires a large amount of computation resources. To overcome this challenge, most current approaches use the ``Autoregressive over X'' architecture, where ``X'' denotes any generative models capable of generating short video clips, including Autoregressive Models like Phenaki~\citep{villegasPhenakiVariableLength2022},  TATS~\citep{geLongVideoGeneration2022}, NUWA-Infinity~\citep{wuNUWAInfinityAutoregressiveAutoregressive2022}; Diffusion Models like MCVD~\citep{voletiMaskedConditionalVideo2022}, FDM~\citep{harveyFlexibleDiffusionModeling2022}, LVDM~\citep{heLatentVideoDiffusion2022}. The main idea behind these approaches is to train the model on short video clips and then use it to generate long videos by a sliding window during inference. ``Autoregressive over X'' architecture not only greatly reduces the computational burden, but also relaxes the data requirements for long videos, as only short videos are needed for training. 

Unfortunately, the ``Autoregressive over X'' architecture, while being a resource-sufficient solution to generate long videos, also introduces new challenges: 1) Firstly, training on short videos but forcing it to infer long videos leads to an enormous training-inference gap. It can result in unrealistic shot change and long-term incoherence in generated long videos, since the model has no opportunity to learn such patterns from long videos. 
For example, Phenaki~\citep{villegasPhenakiVariableLength2022} and TATS~\citep{geLongVideoGeneration2022} are trained on less than 16 frames, while generating as many as 1024 frames when applied to long video generation.
2) Secondly, due to the dependency limitation of the sliding window, the inference process can not be done in parallel and thus takes a much longer time. For example, TATS~\citep{geLongVideoGeneration2022} takes 7.5 minutes to generate 1024 frames, while Phenaki~\citep{villegasPhenakiVariableLength2022} takes 4.1 minutes.

To address the above issues, we propose NUWA-XL, a ``Diffusion over Diffusion'' architecture to generate long videos in a ``coarse-to-fine'' process, as shown in Fig.~\ref{fig:nuwa-xl}. In detail, a global diffusion model first generates $L$ keyframes based on $L$ prompts which forms a ``coarse''  storyline of the video. The first local diffusion model is then applied to $L$ prompts and the adjacent keyframes, treated as the first and the last frames, to complete the middle $L-2$ frames resulting in $L + (L-1)\times(L-2)\approx L^2$ ``fine'' frames in total. By iteratively applying the local diffusion to fill in the middle frames, the length of the video will increase exponentially, leading to an extremely long video. For example, NUWA-XL with $m$ depth and $L$ local diffusion length is capable of generating a long video with the size of $O(L^m)$.  The advantages of such a ``coarse-to-fine'' scheme are three-fold: 1) Firstly, such a hierarchical architecture enables the model to train directly on long videos and thus eliminating the training-inference gap; 2) Secondly, it naturally supports parallel inference and thereby can significantly improve the inference speed when generating long videos; 3) Thirdly, as the length of the video can be extended exponentially w.r.t. the depth $m$, our model can be easily extended to longer videos. Our key contributions are listed in the following:
\begin{itemize}
    \item We propose NUWA-XL, a ``Diffusion over Diffusion'' architecture by viewing long video generation as a novel ``coarse-to-fine'' process.
    \item To the best of our knowledge, NUWA-XL is the first model directly trained on long videos (3376 frames), which closes the training-inference gap in long video generation. 
    \item NUWA-XL enables parallel inference, which significantly speeds up long video generation. Concretely, NUWA-XL speeds up inference by  94.26\% when generating 1024 frames.
    \item We build FlintstonesHD, a new dataset to validate the effectiveness of our model and provide a benchmark for long video generation.
\end{itemize}

\section{Related Work}
\paragraph{Image and Short Video Generation} 

Image Generation has made many progresses, auto-regressive methods~\citep{rameshZeroShotTexttoImageGeneration2021,dingCogviewMasteringTexttoimage2021,yuScalingAutoregressiveModels2022,dingCogView2FasterBetter2022} leverage VQVAE to tokenize the images into discrete tokens and use Transformers~\citep{vaswaniAttentionAllYou2017} to model the dependency between tokens.  DDPM~\citep{hoDenoisingDiffusionProbabilistic2020} presents high-quality image synthesis results. LDM~\citep{rombachHighResolutionImageSynthesis2022} performs a diffusion process on latent space, showing significant efficiency and quality improvements. 

Similar advances have been witnessed in video generation, \citep{vondrickGeneratingVideosScene2016,saitoTemporalGenerativeAdversarial2017,panCreateWhatYou2017,liVideoGenerationText2018,tulyakovMocoganDecomposingMotion2018} extend GAN to video generation. Sync-draw~\citep{mittalSyncdrawAutomaticVideo2017} uses a recurrent VAE to automatically generate videos. GODIVA~\citep{wuGODIVAGeneratingOpenDomaIn2021} proposes a three-dimensional sparse attention to map text tokens to video tokens. VideoGPT~\citep{yanVideoGPTVideoGeneration2021} adapts Transformer-based image generation models to video generation with minimal modifications. NUWA~\citep{wuUWAVisualSynthesis2022} with 3D Nearby Attention extends GODIVA~\citep{wuGODIVAGeneratingOpenDomaIn2021} to various generation tasks in a unified representation. Cogvideo~\citep{hongCogVideoLargescalePretraining2022} leverages a frozen T2I model~\citep{dingCogView2FasterBetter2022} by adding additional temporal attention modules. More recently, diffusion methods~\citep{hoVideoDiffusionModels2022,singerMakeAVideoTexttoVideoGeneration2022,hoImagenVideoHigh2022,zhouMagicVideoEfficientVideo2022} have also been applied to video generation. Among them, \citep{yangDiffusionProbabilisticModeling2022} uses an image diffusion model to predict each individual frame within an RNN temporal autoregressive model. VDM~\citep{hoVideoDiffusionModels2022} replaces the typical 2D U-Net for modeling images with a 3D U-Net. Make-a-video~\citep{singerMakeAVideoTexttoVideoGeneration2022} successfully extends a diffusion-based T2I model to T2V without text-video pairs. Imagen Video~\citep{hoImagenVideoHigh2022} leverages a cascade of video diffusion models to text-conditional video generation.


Different from these works, which concentrate on short video generation, we aim to address the challenges associated with long video generation.

\paragraph{Long Video Generation} 


To address the high computational demand in long video generation, most existing works leverage the ``Autoregressive over X'' architecture, where ``X'' denotes any generative models capable of generating short video clips. With ``X'' being an autoregressive model, NUWA-Infinity~\citep{wuNUWAInfinityAutoregressiveAutoregressive2022} introduces auto-regressive over auto-regressive model, with a local autoregressive to generate patches and a global autoregressive to model the consistency between different patches. TATS~\citep{geLongVideoGeneration2022} presents a time-agnostic VQGAN and time-sensitive transformer model, trained only on clips with tens of frames but can infer thousands of frames using a sliding window mechanism. Phenaki~\citep{villegasPhenakiVariableLength2022} with C-ViViT as encoder and MaskGiT~\citep{changMaskgitMaskedGenerative2022} as backbone generates variable-length videos conditioned on a sequence of open domain text prompts. With ``X'' being diffusion models, MCVD~\citep{voletiMaskedConditionalVideo2022} trains the model to solve multiple video generation tasks by randomly and independently masking all the past or future frames. FDM~\citep{harveyFlexibleDiffusionModeling2022} presents a DDPMs-based framework that produces long-duration video completions in a variety of realistic environments. 

Different from existing ``Autoregressive over X'' models trained on short clips, we propose NUWA-XL, a Diffusion over Diffusion model directly trained on long videos to eliminate the training-inference gap. Besides, the diffusion processes in NUWA-XL can be done in parallel to accelerate the inference.

\section{Method}

\begin{figure*}[!ht]
    \centering
    \includegraphics[width=\linewidth]{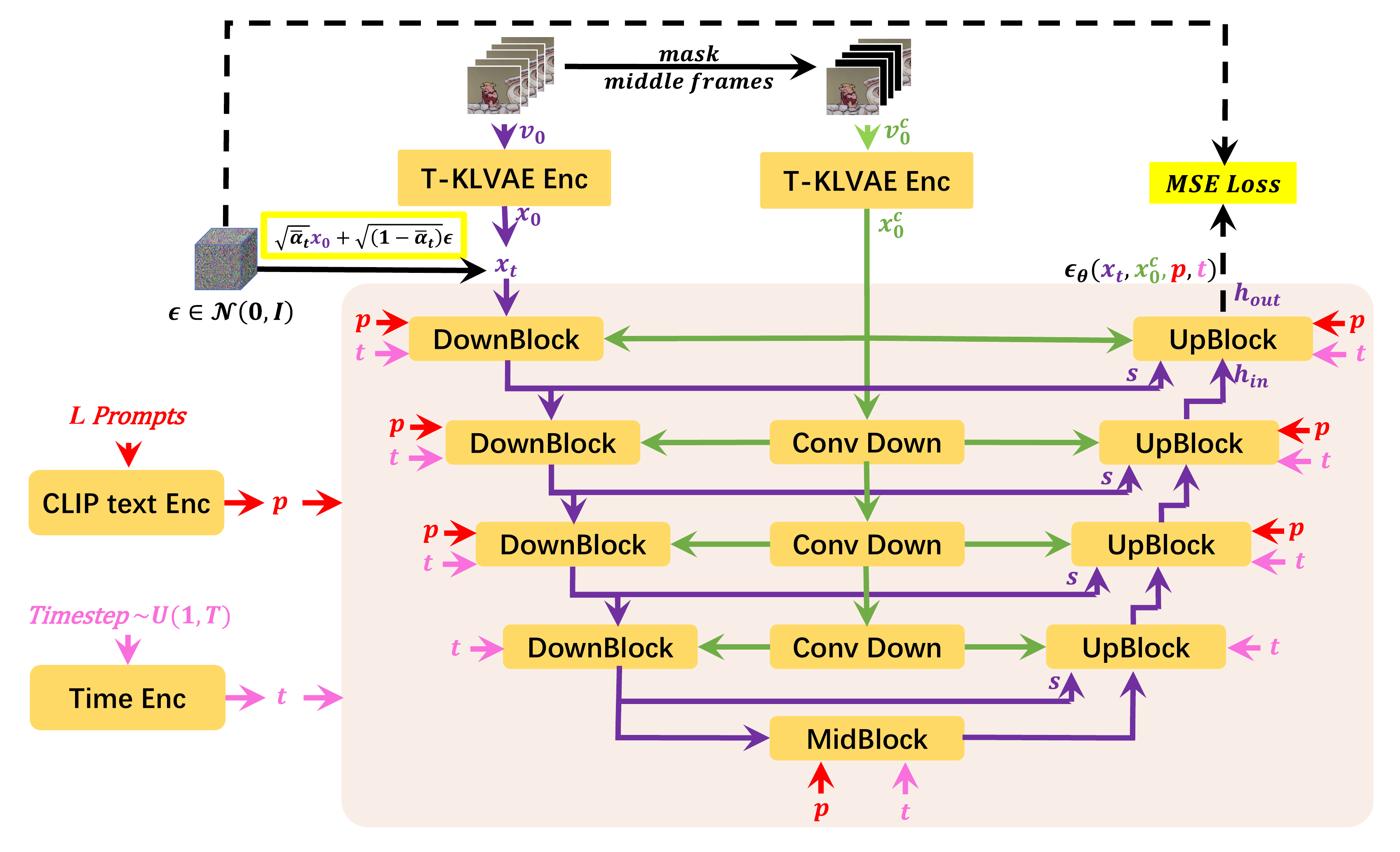}
    \caption{Overview of Mask Temporal Diffusion (MTD) with purple lines standing for diffusion process, red for prompts, pink for timestep, green for visual condition,  black dash for training objective. For global diffusion, all the frames are masked as there are no frames provided as input. For local diffusion,  the middle frames are masked where the first and the last frame are provided as visual conditions. We keep the structure of MTD consistent with the pre-trained text-to-image model as possible to leverage external knowledge.}
    \label{fig:training}
\end{figure*}
\subsection{Temporal KLVAE (T-KLVAE)} \label{sec:T-KLVAE}

Training and sampling diffusion models directly on pixels are computationally costly, KLVAE~\citep{rombachHighResolutionImageSynthesis2022} compresses an original image into a low-dimensional latent representation where the diffusion process can be performed to alleviate this issue. To leverage external knowledge from the pre-trained image KLVAE and transfer it to videos, we propose Temporal KLVAE(T-KLVAE) by adding external temporal convolution and attention layers while keeping the original spatial modules intact. 
 
Given a batch of video $v\in \mathbb{R}^{b\times L\times C\times H\times W}$ with $b$ batch size, $L$ frames, $C$ channels, $H$ height, $W$ width, we first view it as $L$ independent images and encode them with the pre-trained KLVAE spatial convolution. To further model temporal information, we add a temporal convolution after each spatial convolution. To keep the original pre-trained knowledge intact, the temporal convolution is initialized as an identity function which guarantees the output to be exactly the same as the original KLVAE. Concretely, the convolution weight $W^{conv1d}\in \mathbb{R}^{c_{out}\times c_{in}\times k}$ is first set to zero where $c_{out}$ denotes the out channel, $c_{in}$ denotes the in channel and is equal to $c_{out}$, $k$ denotes the temporal kernel size.  Then, for each output channel $i$, the middle of the kernel size $(k - 1)//2$ of the corresponding input channel $i$ is set to 1:
\begin{align}
    W^{conv1d}[i,i,(k - 1)//2] = 1
\end{align}
 Similarly, we add a temporal attention after the original spatial attention, and initialize the weights $W^{att\_out}$ in the out projection layer into zero:
\begin{align} \label{eq:watt}
    W^{att\_out} = 0
\end{align}
For the T-KLVAE decoder $D$, we use the same initialization strategy. The training objective of T-KLVAE is the same as the image KLVAE. Finally , we get a latent code $x_0\in \mathbb{R}^{b\times L\times c\times h\times w}$, a compact representation of the original video $v$.

\subsection{Mask Temporal Diffusion (MTD)}
In this section, we introduce Mask Temporal Diffusion (MTD) as a basic diffusion model for our proposed Diffusion over Diffusion architecture. For global diffusion, only $L$ prompts are used as inputs which form a ``coarse''  storyline of the video, however, for the local diffusion, the inputs consist of not only $L$ prompts but also the first and last frames. Our proposed MTD which can accept input conditions with or without first and last frames, supports both global diffusion and local diffusion. In the following, we first introduce the overall pipeline of MTD and then dive into an UpBlock as an example to introduce how we fuse different input conditions.


Input $L$ prompts, we first encode them by a CLIP Text Encoder to get the prompt embedding $p\in \mathbb{R}^{b\times L\times l_p\times d_p}$ where $b$ is batch size, $l_p$ is the number of tokens, $d_p$ is the prompt embedding dimension. The randomly sampled diffusion timestep $t\sim U(1, T)$ is embedded to timestep embedding $t\in \mathbb{R}^c$. The video $v_0\in\mathbb{R}^{b\times L\times C\times H \times W} $ with $L$ frames is encoded by T-KLVAE to get a representation $x_0\in \mathbb{R}^{b\times L\times c\times h\times w}$. According to the predefined diffusion process:
\begin{align} \label{equ:diffusion}
q\left(x_t\middle| x_{t-1}\right)=\mathcal{N}\left(x_t;\sqrt{\alpha_t}\ x_{t-1},\ \left(1-\alpha_t\right)\mathbf{I}\right) 
\end{align}
$x_0$ is corrupted by:
\begin{align}
    x_t=\sqrt{{\bar{\alpha}}_t}\ x_0+(1-{\bar{\alpha}}_t)\epsilon\quad \epsilon\sim\mathcal{N}(\mathbf{0},\mathbf{I})
\end{align}
where $\epsilon\in \mathbb{R}^{b\times L\times c\times h\times w}$ is noise, $x_t\in \mathbb{R}^{b\times L\times c\times h\times w}$ is the $t$-th intermediate state in diffusion process, $\alpha_t, {\bar{\alpha}}_t$ is hyperparameters in diffusion model.



For the global diffusion model, the visual conditions $v_0^c$ are all-zero. However, for the local diffusion models, $v_0^c\in \mathbb{R}^{b\times L\times C\times H\times W}$ are obtained by masking the middle $L-2$ frames in $v_0$. $v_0^c$ is also encoded by T-KLVAE to get a representation $x_0^c\in \mathbb{R}^{b\times L\times c\times h\times w}$. Finally, the $x_t$, $p$, $t$, $x_0^c$ are fed into a Mask 3D-UNet $\epsilon_\theta\left(\cdot\right)$. Then, the model is trained to minimize the distance between the output of the Mask 3D-UNet $\epsilon_\theta\left(x_t,p,t,x_0^c\right)\in \mathbb{R}^{b\times L\times c\times h\times w}$ and $\epsilon$.
\begin{align}
    \mathcal{L}_\theta=\left|\left|\epsilon-\epsilon_\theta\left(x_t,p,t,x_0^c\right)\right|\right|_2^2
\end{align}

The Mask 3D-UNet is composed of multi-Scale DownBlocks and UpBlocks with skip connection, while the $x_0^c$ is downsampled to the corresponding resolution with a cascade of convolution layers and fed to the corresponding DownBlock and UpBlock.

\begin{wrapfigure}{r}{0.6\textwidth}
    \includegraphics[width=1\textwidth]{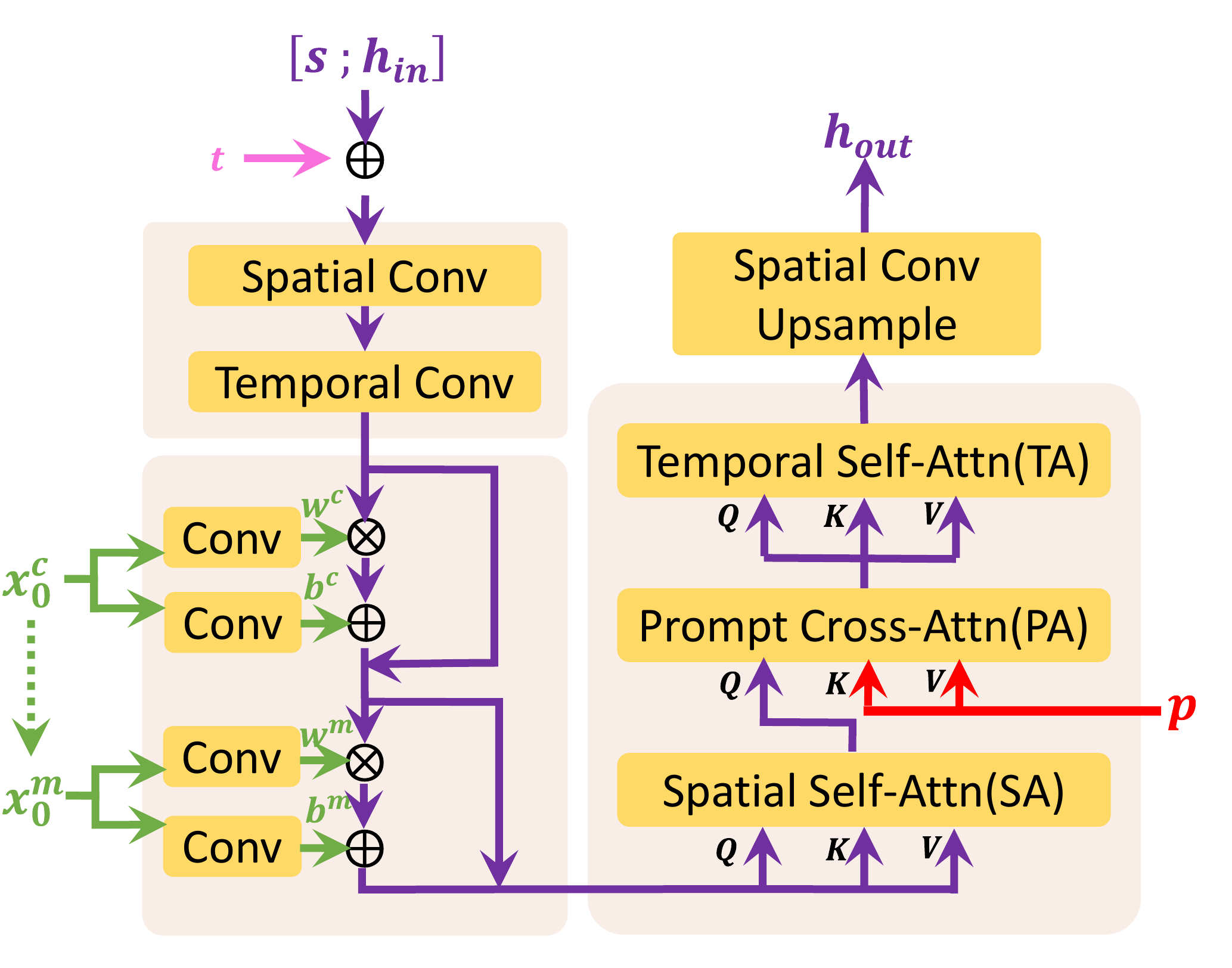}
    \caption{Visualization of the last UpBlock in Mask 3D-UNet with purple lines standing for diffusion process, red for prompts, pink for timestep, green for visual condition.}
    \label{fig:UpBlock}
\end{wrapfigure}

To better understand how Mask 3D-UNet works, we dive into the last UpBlock and show  the details in Fig.~\ref{fig:UpBlock}. The UpBlock takes hidden states $h_{in}$, skip connection $s$, timestep embedding $t$, visual condition $x_0^c$ and prompts embedding $p$ as inputs and output hidden state $h_{out}$. It is noteworthy that for global diffusion, $x_0^c$ does not contain valid information as there are no frames provided as conditions, however, for local diffusion, $x_0^c$ contains encoded information from the first and last frames.

 The input skip connection $s\in \mathbb{R}^{b\times L\times c_{skip}\times h\times w}$ is first concatenated to the input hidden state $h_{in} \in \mathbb{R}^{b\times L\times c_{in}\times h\times w}$.
\begin{align}
    h:=[s;h_{in}]
\end{align}
where the hidden state $h\in \mathbb{R}^{b\times L\times(c_{skip}+c_{in})\times h\times w}$ is then convoluted to target number of channels $h\in \mathbb{R}^{b\times L\times c\times h\times w}$. The timestep embedding $t\in \mathbb{R}^c$ is then added to $h$ in channel dimension $c$.
\begin{align}
    h:=h+t
\end{align}

Similar to Sec.~\ref{sec:T-KLVAE}, to leverage external knowledge from the pre-trained text-to-image model, factorized convolution and attention are introduced with spatial layers initialized from pre-trained weights and temporal layers initialized as an identity function.

For spatial convolution, the length dimension $L$ here is treated as batch-size $h\in \mathbb{R}^{(b \times L)\times c\times h\times w}$. For temporal convolution, the hidden state is reshaped to $h\in \mathbb{R}^{(b \times hw)\times c\times L}$ with spatial axis $hw$ treated as batch-size.
\begin{align}
    h&:=Spatial Conv\left(h\right)\\
    h&:=Temporal Conv\left(h\right)
\end{align}


Then, $h$ is conditioned on $x_0^c\in \mathbb{R}^{b \times L\times c\times h\times w}$ and $x_0^m\in \mathbb{R}^{b\times L\times1\times h\times w}$ where $x_0^m$ is a binary mask to indicate which frames are treated as conditions. They are first transferred to scale $w^c,w^m$ and shift $b^c,b^m$ via zero-initialized convolution layers and then injected to $h$ via linear projection.
\begin{align}
    h&:=w^c\cdot h+b^c+h \\
    h&:=w^m\cdot h+b^m+h
\end{align}

After that, a stack of Spatial Self-Attention (SA), Prompt Cross-Attention (PA), and Temporal Self-Attention (TA) are applied to $h$. 

For the Spatial Self-Attention (SA), the hidden state $h\in \mathbb{R}^{b\times L\times c\times h\times w}$ is reshaped to $h\in \mathbb{R}^{\left(b\times L\right)\times h w\times c}$ with length dimension $L$ treated as batch-size.
\begin{align}
    Q^{SA} &=hW_{Q}^{SA};
    K^{SA} =hW_{K}^{SA};
    V^{SA} =hW_{V}^{SA} \\
    {\widetilde{Q}}^{SA} &=Selfattn(Q^{SA},K^{SA},V^{SA})
\end{align}


where $W_{Q}^{SA}, W_{K}^{SA}, W_{V}^{SA} \in \mathbb{R}^{c\times d_{in}}$ are parameters to be learned. 


For the Prompt Cross-Attention (PA), the prompt embedding $p\in \mathbb{R}^{b\times L\times l_p\times d_p}$ is reshaped to $p\in \mathbb{R}^{\left(b\times L\right)\times l_p\times d_p}$ with length dimension $L$ treated as batch-size.
\begin{align}
    Q^{PA} &=hW_{Q}^{PA};
    K^{PA} =pW_{K}^{PA};
    V^{PA} =pW_{V}^{PA} \\
    {\widetilde{Q}}^{PA} &=Crossattn(Q^{PA},K^{PA},V^{PA})
\end{align}
where $Q^{PA}\in \mathbb{R}^{\left(b\times L\right)\times hw\times d_{in}}$, ${K}^{PA}\in \mathbb{R}^{\left(b\times L\right)\times l_p\times d_{in}}$, $V^{PA}\in \mathbb{R}^{\left(b\times L\right)\times l_p\times d_{in}}$ are query, key and value, respectively. $W_{Q}^{PA}\in \mathbb{R}^{c\times d_{in}}$, $W_{K}^{PA}\in \mathbb{R}^{d_p\times d_{in}}$ and $W_{V}^{PA}\in \mathbb{R}^{d_p\times d_{in}}$ are parameters to be learned. 




The Temporal Self-Attention (TA) is exactly the same as Spatial Self-Attention (SA) except that spatial axis $hw$ is treated as batch-size and temporal length $L$ is treated as sequence length.

Finally, the hidden state $h$ is upsampled to target resolution $h_{out}\in \mathbb{R}^{b\times L\times c\times h_{out}\times w_{out}} $ via spatial convolution.
Similarly, other blocks in Mask 3D-UNet leverage the same structure to deal with the corresponding inputs.

\subsection{Diffusion over Diffusion Architecture}
In the following, we first introduce the inference process of MTD, then we illustrate how to generate a long video via Diffusion over Diffusion Architecture in a novel ``coarse-to-fine'' process.

In inference phase, given the $L$ prompts $p$ and visual condition $v_0^c$,  $x_0$ is sampled from a pure noise $x_T$ by MTD. Concretely, for each timestep $t=T,T-1,\ldots,1$, the intermediate state $x_t$ in diffusion process is updated by
\begin{align}
    x_{t-1}=\frac{1}{\sqrt{\alpha_t}}\left(x_t-\frac{1-\alpha_t}{\sqrt{\left(1-{\bar{\alpha}}_t\right)}}\epsilon_\theta\left(x_t,p,t,x_0^c\right)\right)+\frac{\left(1-{\bar{\alpha}}_{t-1}\right)\beta_t}{1-{\bar{\alpha}}_t}\cdot \epsilon 
\end{align}
where $\epsilon\sim\mathcal{N}(\mathbf{0},\mathbf{I})$, $p$ and $t$ are embedded prompts and timestep, $x_0^c$ is encoded $v_0^c$. $\alpha_t$, ${\bar{\alpha}}_t$, $\beta_t$ are hyperparameters in MTD.

Finally, the sampled latent code $x_0$ will be decoded to video pixels $v_0$ by T-KLVAE. For simplicity, the iterative generation process of MTD is noted as 
\begin{align}
    v_0=Diffusion(p, v_0^c)
\end{align}

When generating long videos, given the $L$ prompts $p_1$ with large intervals, the $L$ keyframes are first generated through a global diffusion model.
\begin{align}
    v_{01}=Global Diffusion(p_1,v_{01}^c)
\end{align}
where $v_{01}^c$ is all-zero as there are no frames provided as visual conditions. The temporally sparse keyframes $v_{01}$ form the ``coarse'' storyline of the video. 

Then, the adjacent keyframes in $v_{01}$ are treated as the first and the last frames in visual condition $v_{02}^c$. The middle $L-2$ frames are generated by feeding $p_2$, $v_{02}^c$ into the first local diffusion model where $p_2$ are $L$ prompts with smaller time intervals.  
\begin{align}
    v_{02}=Local Diffusion(p_2,v_{02}^c)
\end{align}

Similarly, $v_{03}^c$ is obtained from adjacent frames in $v_{02}$, $p_3$ are $L$ prompts with even smaller time intervals. The $p_3$ and $v_{03}^c$ are fed into the second local diffusion model.
\begin{align}
    v_{03}=Local Diffusion(p_3,v_{03}^c)
\end{align}
Compared to frames in $v_{01}$, the frames in $v_{02}$ and $v_{03}$ are increasingly ``fine'' with stronger consistency and more details.

By iteratively applying the local diffusion to complete the middle frames, our model with $m$ depth is capable of generating extremely long video with the length of $O(L^m)$. Meanwhile, such a hierarchical architecture enables us to directly train on temporally sparsely sampled frames in long videos (3376 frames) to eliminate the training-inference gap. After sampling the $L$ keyframes by global diffusion, the local diffusions can be performed in parallel to accelerate the inference speed.

\section{Experiments}


\subsection{FlintstonesHD Dataset}
Existing annotated video datasets have greatly promoted the development of video generation. However, the current video datasets still pose a great challenge to long video generation. First, the length of these videos is relatively short, and there is an enormous distribution gap between short videos and long videos such as shot change and long-term dependency. Second, the relatively low resolution limits the quality of the generated video. Third, most of the annotations are coarse descriptions of the content of the video clips, and it is difficult to illustrate the details of the movement.

 To address the above issues, we build FlintstonesHD dataset, a densely annotated long video dataset, providing a benchmark for long video generation. We first obtain the original \textit{Flintstones} cartoon which contains 166 episodes with an average of 38000 frames of $1440\times 1080$ resolution. To support long video generation based on the story and capture the details of the movement, we leverage the image captioning model GIT2~\citep{wangGITGenerativeImagetotext2022} to generate dense captions for each frame in the dataset first and manually filter some errors in the generated results. 

 \subsection{Metrics}
\paragraph{Avg-FID} Fréchet Inception Distance(FID)~\citep{heuselGansTrainedTwo2017}, a metric used to evaluate image generation, is introduced to calculate the average quality of generated frames.

\paragraph{Block-FVD}  Fréchet Video Distance (FVD)~\citep{unterthinerAccurateGenerativeModels2018} is widely used to evaluate the quality of the generated video. In this paper, we propose Block FVD for long video generation, which splits a long video into several short clips to calculate the average FVD of all clips. For simplicity, we name it B-FVD-X where X denotes the length of the short clips.


\subsection{Quantitative Results}
\begin{table*}[!ht]
\newcommand{\bwidth}{2.4cm}
\newcommand{\bbwidth}{3.0cm}
\begin{tabular} {p{0.8cm}p{1.9cm}p{2.2cm}p{2.2cm}p{2.2cm}p{2.2cm}}
\toprule
\multicolumn{2}{c}{Method}  & Phenaki~\citep{villegasPhenakiVariableLength2022} & FDM*~\citep{harveyFlexibleDiffusionModeling2022} &   NUWA-XL & NUWA-XL      \\ 
\midrule
\multicolumn{2}{c}{Resolution}    & 128      & 128  &   128  & 256      \\ 
\midrule
\multicolumn{2}{c}{Arch}    & AR over AR      & AR over Diff  &   Diff over Diff  & Diff over Diff       \\ 
\midrule
\multirow{3}{*}{16f}    & Avg-FID↓        &       40.14      &  34.47        &    35.95   &     32.66       \\
                        & B-FVD-16↓   &       544.72     &  532.94       &    520.19  &  580.21         \\
                        & Time↓       &         4s         &        7s       &    7s    &   15s       \\ 
\midrule
\multirow{3}{*}{256f}   & Avg-FID↓        &       43.13     &     38.28 &     35.68     &  32.05       \\
                        & B-FVD-16↓   &        573.55    &    561.75         &     542.26    &   609.32   \\
                        & Time↓       &      65s      &     114s        &       17s \textcolor[rgb]{0,0.391,0}{(85.09\%↓)}   &   32s   \\ 
\midrule
\multirow{3}{*}{1024f}  & Avg-FID↓        &       48.56     &     43.24        &       35.79    &     32.07          \\
                        & B-FVD-16↓   &       622.06    &     618.42       &        572.86   &     642.87\\
                        & Time↓       &       259s      &     453s        &       26s \textcolor[rgb]{0,0.391,0}{(94.26\%↓)}  &   51s  \\ 
\bottomrule
\end{tabular}
\caption{Quantitative comparison with the state-of-the-art models for long video generation on FlintstonesHD dataset. 128 and 256 denote the resolutions of the generated videos. *Note that the original FDM model does not support text input. For a fair comparison, we implement an FDM with text input. }\label{tab:long_video}

\end{table*}
\subsubsection{Comparison with the state-of-the-arts}
We compare NUWA-XL on FlintstonesHD with the state-of-the-art models in Tab.~\ref{tab:long_video}. Here, we report FID, B-FVD-16, and inference time. For ``Autoregressive over X (AR over X)'' architecture, due to error accumulation, the average quality of generated frames (Avg-FID)
declines as the video length increases. However, for NUWA-XL, where the frames are not generated sequentially, the quality does not decline with video length. Meanwhile, compared to ``AR over X'' which is trained only on short videos, NUWA-XL is capable of generating higher quality long videos. As the video length grows, the quality of generated segments (B-FVD-16) of NUWA-XL declines more slowly as NUWA-XL has learned the patterns of long videos. Besides, because of parallelization,  NUWA-XL significantly improves the inference speed by 85.09\% when generating 256 frames and by 94.26\% when generating 1024 frames.

\subsubsection{Ablation study}

\begin{table*}[!ht]
\newcommand{\bwidth}{0.6cm}
\newcommand{\bbwidth}{0.8cm}
\begin{subtable}{0.48\linewidth}
\centering 
\begin{tabular}{p{1.8cm}p{2.4cm}p{0.4cm}p{0.5cm}}
\toprule
Model   & Temporal Layers & FID↓ & FVD↓ \\
\midrule
KLVAE   &    -        &   4.71   &  28.07    \\
T-KLVAE-R & random init &   5.44   &  12.75    \\
T-KLVAE & identity init &   4.35   &  11.88   \\
\bottomrule
\end{tabular}
\caption{Comparison of different KLVAE settings.}\label{tab:KLVAE}
\end{subtable}
\begin{subtable}{0.48\textwidth}
\centering 
 \begin{tabular}{p{2.0cm}p{0.3cm}p{0.3cm}p{0.5cm}p{\bbwidth}}
\toprule
Model & MI & SI  & FID↓ & FVD↓ \\
\midrule
MTD w/o MS   &  × & × &  39.28  &   548.90   \\
MTD w/o S    &   \checkmark & × &  36.04 &  526.36    \\
MTD         &   \checkmark & \checkmark & 35.95   &   520.19  \\
\bottomrule
\end{tabular}
\caption{Comparison of different MTD settings.}\label{tab:MTD}
\end{subtable}

\begin{subtable}{0.48\linewidth}
\centering
\begin{tabular}{p{2.2cm}p{0.45cm}p{0.65cm}p{0.65cm}p{0.65cm}}
\toprule
Model      & depth & 16f & 256f & 1024f \\
\midrule
NUWA-XL-D1 & 1     & 527.44 & 697.20  &  719.23    \\
NUWA-XL-D2 & 2     & 516.05 & 536.98  &  684.57   \\
NUWA-XL-D3 & 3     & 520.19 & 542.26  &  572.86  \\
\bottomrule
\end{tabular}
\caption{Comparison of different NUWA-XL depth.}\label{tab:depth}
\end{subtable}
\begin{subtable}{0.48\linewidth}
\centering
\begin{tabular}{p{2.4cm}p{0.2cm}p{0.65cm}p{0.65cm}p{0.65cm}}
\toprule
Model       & L  & 16f & 256f & 1024f \\
\midrule
NUWA-XL-L8  & 8  & 569.43 & 673.87  &  727.22     \\
NUWA-XL-L16 & 16 & 520.19 & 542.26  &  572.86  \\
NUWA-XL-L32 & 32 & OOM & OOM  & OOM  \\
\bottomrule
\end{tabular}
\caption{Comparison of different local diffusion length. }\label{tab:length}
\end{subtable}
\caption{Ablation experiments for long video generation on FlintstonesHD (OOM stands for Out Of Memory).} \label{tab:ablation}

\end{table*}

\paragraph{KLVAE} Tab.~\ref{tab:KLVAE} shows the comparison of different KLVAE settings. KLVAE means treating the video as independent images and reconstructing them independently. T-KLVAE-R means the introduced temporal layers are randomly initialized. Compared to KLVAE, we find the newly introduced temporal layers can significantly increase the ability of video reconstruction.  Compared to T-KLVAE-R, the slightly better FID and FVD in T-KLVAE illustrate the effectiveness of identity initialization.

\paragraph{MTD} Tab.~\ref{tab:MTD} shows the comparison of different global/local diffusion settings. MI (Multi-scale Injection) means whether visual conditions are injected to multi-scale DownBlocks and UpBlocks in Mask 3D-UNet or only injected to  the Downblock and UpBlock with the highest scale. SI (Symmetry Injection) means whether the visual condition is injected into both DownBlocks and UpBlocks or it is only injected into UpBlocks. Comparing MTD w/o MS and MTD w/o S, multi-scale injection is significant for long video generation. Compared to MTD w/o S, the slightly better FID and FVD in MTD show the effectiveness of symmetry injection.

\paragraph{Depth of Diffusion over Diffusion} 

Tab.~\ref{tab:depth} shows the comparison of B-FVD-16 of different NUWA-XL depth $m$ with local diffusion length $L$ fixed to 16. When generating 16 frames, NUWA-XL with different depths achieves comparable results. However, as the depth increases, NUWA-XL can produce videos that are increasingly longer while still maintaining relatively high quality.

\paragraph{Length in Diffusion over Diffusion}  Tab.~\ref{tab:length} shows the comparison of B-FVD-16 of diffusion local length $L$ with NUWA-XL depth $m$ fixed to 3. In comparison, when generating videos with the same length, as the local diffusion length increases, NUWA-XL can generate higher-quality videos.

\subsection{Qualitative results}
\begin{figure*}[!ht]
    \centering
    \includegraphics[width=\linewidth]{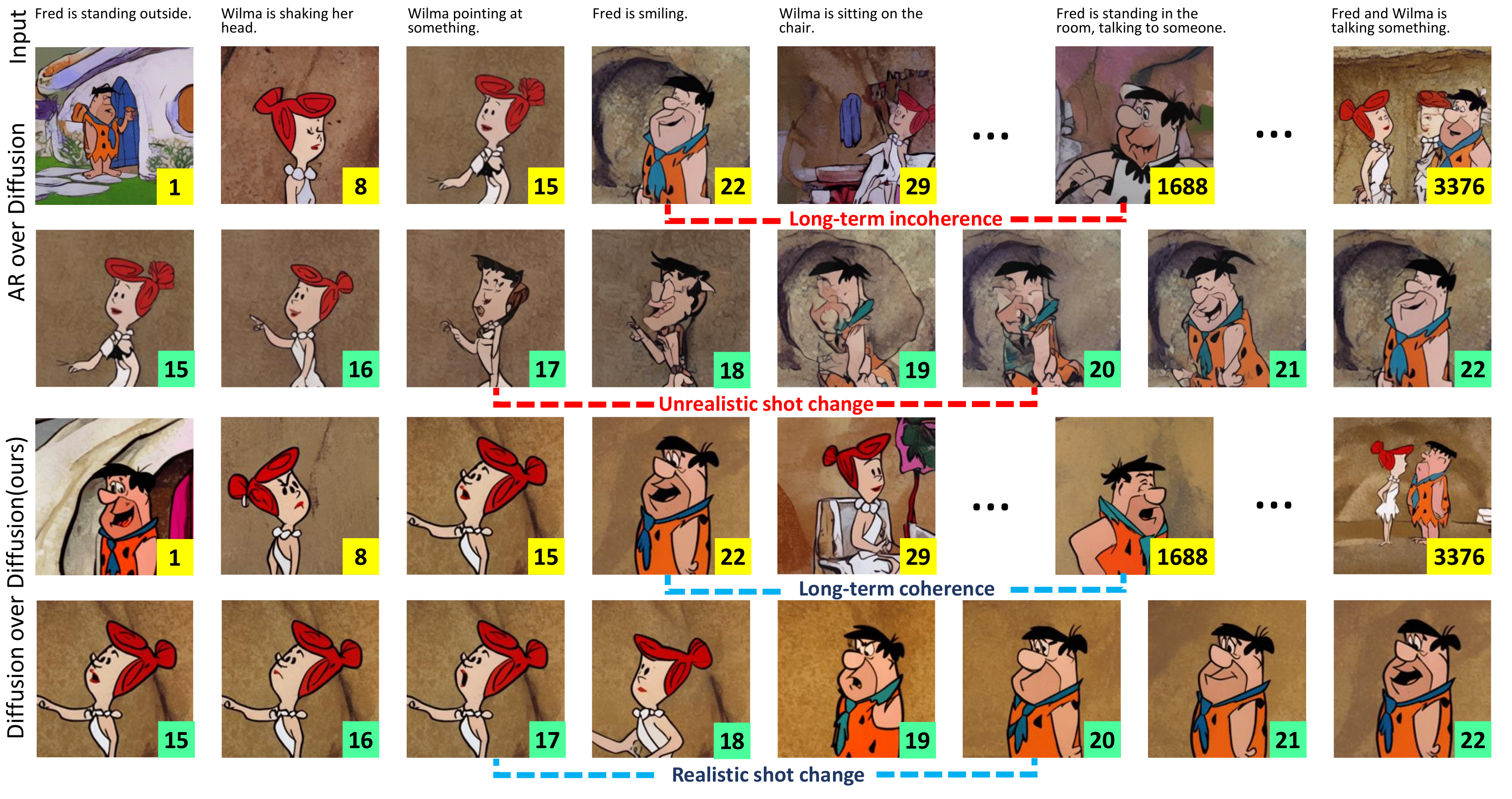}
    \caption{Qualitative comparison between AR over Diffusion and Diffusion over Diffusion for long video generation on FlintstonesHD. The Arabic number in the lower right corner indicates the frame number with yellow standing for keyframes with large intervals and green for small intervals. Compared to AR over Diffusion, NUWA-XL generates long videos with long-term coherence (see the cloth in frame 22 and 1688) and realistic shot change (frame 17-20).} 
    \label{fig:comparison}
\end{figure*}
Fig.~\ref{fig:comparison} provides a qualitative comparison between AR over Diffusion and Diffusion over Diffusion for long video generation on FlintstonesHD. As introduced in Sec.~\ref{sec:intro}, when generating long videos,  ``Autoregressive over X'' architecture trained only on short videos will lead to long-term incoherence (between frame 22 and frame 1688) and unrealistic shot change (from frame 17 to frame 20) since the model has no opportunity to learn the distribution of long videos. However, by training directly on long videos, NUWA-XL successfully models the distribution of long videos and generates long videos with long-term coherence and realistic shot change. 

\section{Conclusion}

We propose NUWA-XL, a ``Diffusion over Diffusion'' architecture by viewing long video generation as a novel ``coarse-to-fine'' process. To the best of our knowledge, NUWA-XL is the first model directly trained on long videos (3376 frames), closing the training-inference gap in long video generation. Additionally, NUWA-XL allows for parallel inference, greatly increasing the speed of long video generation by 94.26\% when generating 1024 frames. We further build FlintstonesHD, a new dataset to validate the effectiveness of our model and provide a benchmark for long video generation.

\section{Limitations}

Although our proposed NUWA-XL improves the quality of long video generation and accelerates the inference speed, there are still several limitations: First, due to the unavailability of open-domain long videos (such as movies, and TV shows), we only validate the effectiveness of NUWA-XL on public available cartoon Flintstones. We are actively building an open-domain long video dataset and have achieved some phased results, we plan to extend NUWA-XL to open-domain in future work. Second, direct training on long videos reduces the training-inference gap but poses a great challenge to data. Third, although NUWA-XL can accelerate the inference speed, this part of the gain requires reasonable GPU resources to support parallel inference. 

\section{Ethics Statement}



This research is done in alignment with Microsoft’s responsible AI principles.

\section*{Acknowledgements}
We'd like to thank Yu Liu, Jieyu Xiao, and Scarlett Li for discussion of the potential cartoon scenarios. We'd also like to thank Yang Ou and Bella Guo for the design of the homepage. We's also like to thank Yan Xia, Ting Song, and Tiantian Xue for the implementation of the homepage. 

\bibliographystyle{plain}
\bibliography{nuwaxl}

\end{document}